\documentclass[conference]{IEEEtran}
\IEEEoverridecommandlockouts
\usepackage{cite}
\usepackage{amsmath,amssymb,amsfonts}
\usepackage{algorithmic}
\usepackage{graphicx}
\usepackage{textcomp}
\usepackage{xcolor}
\def\BibTeX{{\rm B\kern-.05em{\sc i\kern-.025em b}\kern-.08em
    T\kern-.1667em\lower.7ex\hbox{E}\kern-.125emX}}
    
\usepackage{caption}
\begin{document}

\title{Computer Vision based Accident Detection for Autonomous Vehicles}


\author
{
\IEEEauthorblockN{Dhananjai Chand \IEEEauthorrefmark{1}, Savyasachi Gupta \IEEEauthorrefmark{2},
Ilaiah Kavati \IEEEauthorrefmark{3}}
\IEEEauthorblockA{Department of Computer Science and Engineering\\
National Institute of Technology Warangal, India-506004\\
Email : \IEEEauthorrefmark{1} dcdevdhan@gmail.com, \IEEEauthorrefmark{2} gsavya10@gmail.com,
\IEEEauthorrefmark{3}  ilaiahkavati@nitw.ac.in\\
ORCID : \IEEEauthorrefmark{1} 0000-0003-2870-1514, \IEEEauthorrefmark{2} 0000-0002-9080-2028,
\IEEEauthorrefmark{3} 0000-0002-2659-2329}
}

\IEEEpubid{\makebox[\columnwidth]{978-1-7281-6916-3/20/\$31.00~\copyright2020 IEEE \hfill} \hspace{\columnsep}\makebox[\columnwidth]{ }}

\maketitle

\begin{abstract}
Numerous Deep Learning and sensor-based models have been developed to detect potential accidents with an autonomous vehicle. However, a self-driving car needs to be able to detect accidents between other vehicles in its path and take appropriate actions such as to slow down or stop and inform the concerned authorities. In this paper, we propose a novel support system for self-driving cars that detects vehicular accidents through a dashboard camera. The system leverages the Mask R-CNN framework for vehicle detection and a centroid tracking algorithm to track the detected vehicle. Additionally, the framework calculates various parameters such as speed, acceleration, and trajectory to determine whether an accident has occurred between any of the tracked vehicles. The framework has been tested on a custom dataset of dashcam footage and achieves a high accident detection rate while maintaining a low false alarm rate.
\end{abstract}


\section{Introduction}
Road-side traffic flow has turned out to be an elementary share of human lifestyles and has an impact on several services and activities on a day-to-day basis. It can be seen that 1.25 million human beings lose their lives due to vehicular accidents annually with a further 20 to 50 million people suffering injuries and disabilities \cite{SG_14_6}. It is reported that the number of people who lose their lives and the amount of collateral damage has risen exponentially proportional to the count of road accidents and the number of vehicles manufactured on an annual basis\cite{SG_14_3}. Even though several steps are continuously taken to improve CCTV surveillance by the placement of cameras at road junctions \cite{SG_14_4} and high-tech radars for capturing over-speeding vehicles on expressways \cite{SG_14_5,SG_14_7,SG_14_8}, several human casualties occur due to delays in reporting accident cases in a timely fashion \cite{SG_14_3} causing further delays for receiving medical assistance.\par
To alleviate the given problem, this work proposes a solution which addresses the given problem by detecting and reporting vehicular collisions almost spontaneously, which can help local medical authorities and traffic police to address the condition in a timely manner. \par

\section{Related Work}
\label{section1}
The use of computer vision has garnered a lot of interest and promise in the areas of traffic surveillance and vehicular accident detection. For instance, we can use accident detection to alert the autonomous vehicle to take caution or stop itself in the presence of an accident. Secondly, autonomous vehicles can report the local authorities when they detect an already occurred accident which can speed up the process of alerting the medics and handling the incidents post-accidents.\par 
Hui \emph{et al.}\cite{SG_24_1} introduced a solution which utilizes Gaussian Mixture Model (GMM) to discover vehicles. These vehicles, which have been discovered, are tracked using a mean shift algorithm. However, this approach does not perform well in scenarios where erratic changes are present in the traffic pattern or the case of acute weather conditions since it relies on a limited number of parameters. Another approach was established by Singh \emph{et al.}\cite{SG_24_2} for determining vehicular accidents based on the visuals of a traffic surveillance camera by creating Spatio-Temporal Video Volumes (STVVs) and then obtaining deep representations on denoising autoencoders. These are then used to create an anomaly score, as well as at the same time detect and track moving objects and calculate the probability of an accident by determining the intersection of the movements of the vehicles. However, this approach works well with normal traffic flow at junctions under well-lit conditions but its performance degrades severely for accurately predicting and detecting accidents in the case of variations in traffic flow, occlusions of car accidents, and poor visibility conditions.\par
Hence, another method is needed which could work reliably on any footage under any condition, which is illustrated in the methodology section.\par

\section{Proposed Approach}
\label{accident_method}
\begin{figure}[!ht]
    \centering
	\includegraphics[width=.95\linewidth]{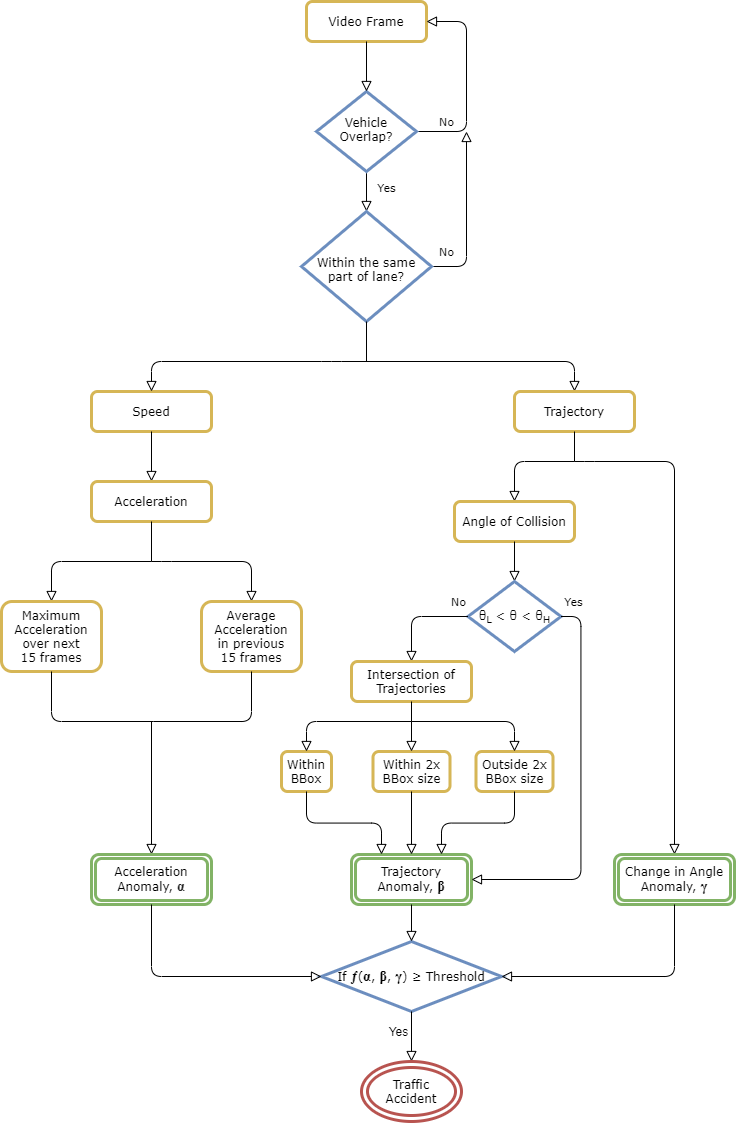}
	\caption{Workflow diagram depicting the sequence of actions taken for vehicular accident detection. }
	\label{fig:accidentworkflow}
\end{figure}

The frequency of vehicular accidents has steadily risen and accident detection using computer vision is a valuable, but challenging task today. In the following section, a framework for the detection of vehicular accidents is proposed. This framework uses Mask R-CNN \cite{DC_37_1} for object detection and an object tracking algorithm \cite{SG_43_1} on dashcam videos. The chances of an accident happening are decided based on a set of anomalies that are determined in a vehicle when it overlaps with another vehicle in the footage. The following section depicts our proposed system illustrated in \figurename~\ref{fig:accidentworkflow}. A key objective is to give a basic yet efficient method for solving the problem of vehicular accident detection which can work productively and give crucial data to paramedical services and concerned authorities immediately. The outlined model is an extension of our research conducted in a similar domain but on CCTV camera footage \cite{SG_44_1}. Change of angle of view in dashcam poses a new set of challenges and we have modified our approach tackling the same.\par
The following sections explain our approach in detail.

\subsection{Vehicle Detection}
This section explains how vehicles are identified in the input video. The primary object detection model used for vehicle detection is Mask R-CNN\cite{DC_37_1} which is used for identifying vehicles namely in the following categories: car, bus, and truck. One of the key reasons for using Mask R-CNN for object detection is the fact that its weights after training on the MS COCO dataset \cite{mscoco} are easily accessible. Since the objective of this support system is to identify vehicles on Indian roads, we evaluated the ability of the model by testing on dashcam footage containing conditions such as varying degree of traffic, weather, and lighting conditions.\par
The outcome of this step is an output dictionary from the use of the Mask R-CNN model, which contains the class IDs, generated masks, bounding boxes, and the detection scores of vehicles for a video frame.

\subsection{Vehicle Tracking}
In the next step, correctly detected vehicles are retained based on their class IDs and scores by filtering from all detected objects. This paves the way for the next task which is vehicle tracking over the recording's future frames. A centroid tracking \cite{SG_43_1} algorithm is used to achieve the task of object tracking, which is a simple yet efficient method. This algorithm functions by finding the Euclidean distance between the detected vehicles' centroids over successive frames as clarified further ahead. The centroid tracking algorithm used in this framework can be described as a multiple-step technique which is explained in the following steps:
\begin{enumerate}
	\item The object centroid for all objects is determined by taking the midpoint of the meeting lines from the middle of the bounding boxes for each distinguished vehicle. 
	\item The Euclidean distance is determined between the centroids of the newly recognized objects and the current objects. 
	\item The centroid coordinates for the new items are refreshed depending on the least Euclidean distance from the current set of centroids and the newly stored centroid. 
	\item New objects in the field of view are registered by storing their centroid coordinates in a dictionary and appointing fresh IDs. 
	\item Objects which are not noticeable in the current field of view for a predefined set of frames in progression are deregistered from the dictionary.
\end{enumerate}

\par An important presumption of the centroid tracking algorithm is that though a vehicle will move between the resulting frames of the recording, the separation between the centroids of the same vehicle between successive frames will be less than the separation to the centroid of some other vehicle identified in the frame.

\subsection{Vehicle Overlap}
There are strong chances that the bounding boxes of detected vehicles overlap before a collision occurs between them. Hence, the first step is to determine whether a pair of detected vehicles overlap or not and is explained as follows.
\par Let $a$ and $b$ to be the bounding boxes of two vehicles $A$ and $B$. Then, $x$ and $y$ are taken as the pair of coordinates of the centroid of a given vehicle, and $\alpha$ and $\beta$ represent the width and height of the bounding box of a vehicle respectively. Now, at any instance, if the condition \eqref{eq:1} holds, we determine that the bounding boxes of $A$ and $B$ overlap.
\begin{equation}\label{eq:1}
( 2 \times |a.x - b.x| < a.\alpha + b.\alpha) \land ( 2 \times |a.y + b.y| < a.\beta + b.\beta)
\end{equation}

The equation given above verifies whether the bounding boxes' centres for vehicles $A$ and $B$ are close enough to intersect on both axes. The vehicles are considered to overlap if the bounding boxes are found to intersect on both the axes. This is an essential and preliminary step in the framework which serves as the grounds for checking the other set of conditions as mentioned earlier.

\subsection{Lane Detection and Section-wise Analysis}
\begin{figure}[!ht] 
	\centering
	\includegraphics[width=\linewidth, height=45mm]{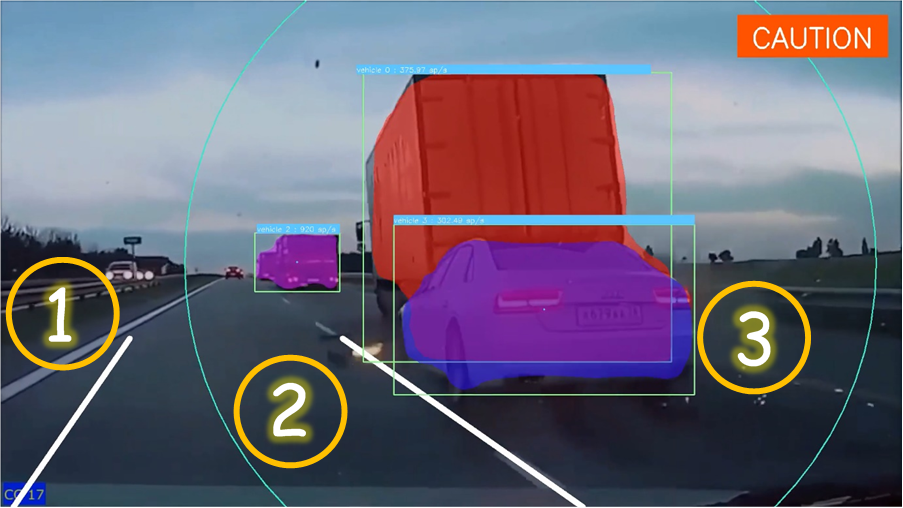}
	\caption{Sections of a Lane}
	\label{fig:lane_segregation}
\end{figure}

We use an efficient lane detection system proposed by Aggarwal \cite{SG_39_1}, to identify lane demarcations on the road. It is used to determine whether the vehicles which have been found to overlap lie within the same side of the lane or not. To understand this, we separate the road into three sections as shown in \figurename~\ref{fig:lane_segregation}. The three sections are as follows:
\begin{itemize}
	\item [S1:] Left side of the `left’ lane demarcation
	\item [S2:] Between both of the lane demarcations
	\item [S3:] Right side of the `right’ lane demarcation
\end{itemize}

These sections are used to fulfill the basic criteria of vehicles involved in accidents. Vehicles that are involving in an accident must have at least half of their bounding box in the same section. This serves as an additional check for vehicle overlaps which is useful for making the framework more robust to the change in the angle of view present in dashcam (i.e. front-view) compared to CCTV surveillance (i.e. isometric view). For instance, \figurename~\ref{fig:lane_segregation} illustrates a case where the vehicles involving in an overlap are both in section S3 and hence the possibility of an accident exists.

\subsection{Vehicle Trajectory and Angle of Intersection determination}
\label{trajectory}
In this step, we determine the trajectories of the vehicles. The trajectory of a tracked vehicle is calculated by finding the Euclidean distance between its centroids in the current frame and five frames ago. This task is achieved by storing the list of centroids of each tracked vehicle in a dictionary as long as the vehicle is registered as per the working of the centroid tracking algorithm. The result is a 2D vector, $\boldsymbol{\nu}$, which represents the movement direction of the vehicle. The size of the vector, $\boldsymbol{\mu}$, is calculated from \eqref{eq:mag}.
\begin{equation}\label{eq:mag}
\text{magnitude}= \sqrt {\left( {\boldsymbol{\mu}.i } \right)^2 + \left( {\boldsymbol{\mu}.j } \right)^2 }
\end{equation}

This vector, $\boldsymbol{\nu}$, is then normalized by dividing it by its magnitude and is subsequently stored in a dictionary consisting of normalized direction vectors of every tracked object only if its original magnitude is over a certain threshold. If not, this vector is discarded. This is done to guarantee that minor variations in centroids for static objects don't bring about a false change of trajectories.\par

Next, the angle between trajectories of overlapping vehicles is calculated by using the equation for finding the angle between the two vectors. Here, $\boldsymbol{\mu_1}$ and $\boldsymbol{\mu_2}$ are viewed as the direction vectors for each of the intersecting vehicles respectively. Subsequently, the angle of intersection between the two directions $\theta$ is determined by the following formula in \eqref{eq:angle}.
\begin{equation}\label{eq:angle}
\theta = \arccos \left(\frac{\boldsymbol{\mu_1 \cdot \mu_2}}{\boldsymbol{|\mu_1| |\mu_2|}}\right)
\end{equation}

The use of the angle of intersection of vehicle trajectories is to serve as a parameter for determining if a collision has occurred. \par

\subsection{Vehicle Speed and Change in Acceleration determination}
In this step, we determine the absolute speed (in metres/second) at which a tracked vehicle is moving. This is needed to calculate further parameters and is computed by finding some intermediate values listed sequentially in the following order:
\begin{enumerate}
	\item The scaled speed (in pixels/second) of tracked vehicles in the frame relative to the speed of the vehicle on which the dashcam is mounted
	\item The absolute speed (in metres/second) of the vehicle on which the dashcam is mounted
	\item The absolute speed (in metres/second) of tracked vehicles in the frame
\end{enumerate}
Once the absolute speeds of the tracked vehicles are determined, we can calculate the acceleration and its change to use it as a parameter whether an accident has occurred or not.

\subsubsection{Scaled speed of tracked vehicles relative to the view of the observing vehicle}
\label{relative_speed}
The relative speed of the vehicle is evaluated in a progression of steps. At first, the gap between the frames of the video, $\tau$, is assessed utilizing the frames per second (FPS) metric as given in \eqref{eq:4}.
\begin{equation}\label{eq:4}
\tau = \frac{1}{\text{FPS}}
\end{equation}
At that point, the distance covered by a vehicle over five frames from the centroid of the vehicle, ${c_1}$, in the first frame and, ${c_2}$, in the fifth frame is calculated. If the vehicle has not been distinguished in the frame for five seconds, the most recently accessible past centroid is taken. The Gross Speed (${S_g}$) is resolved from centroid difference found out over the ${Interval}$ of five frames using \eqref{eq:5}.
\begin{equation}\label{eq:5}
S_g = \frac{c_2 - c_1}{\tau \times Interval}
\end{equation}
Subsequently, the moving velocity of the vehicle is normalized regardless of its separation from the camera utilizing condition \eqref{eq:6} by taking the height of the video frame (${H}$) and the height of the bounding box of the vehicle (${h}$) to acquire the scaled speed (${S_s}$) of the vehicle. The scaled speeds of the tracked vehicles are also kept in a dictionary for each frame.
\begin{equation}\label{eq:6}
S_s = \left(\frac{H - h}{H} + 1\right) \times S_g
\end{equation}
We use this scaled speed (${S_s}$) to calculate the absolute speed of the vehicle in further explanation.

\subsubsection{Absolute speed of the vehicle on which the dashcam is mounted using the optical flow method}
\label{own_absolute_speed}
As the vehicle on which the dashcam is mounted is moving along the road, it will have its own speed and trajectory. It has to be taken into consideration that the speed and trajectory values found in the previous steps are relative to our vehicle. Therefore, we must find the absolute values of speed and trajectory for our vehicle. To compute these two features, we will use a classical computer vision technique, which is optical flow. Optical flow alludes to the apparent movement of objects in the picture because of the relative movement between the camera and the scene. It is determined for specific features on the frame and can be shown as a vector tangent to the movement of those specific features on the frame. One of the well-known procedures of solving for optical flow is the Lucas-Kanade method \cite{SG_44_2} which we will be using for our approach.\par
$OpenCV$ \cite{SG_44_4} allows us to perform this operation directly via an in-built function,\emph{v2.calcOpticalFlowPyrLK()}, to which we feed the old frame and features from the previous frame. Features are chosen using Shi-Tomasi \cite{SG_44_3} corner detection algorithm which is also in-built in $OpenCV$. Let there be $n$ total features that are detected. We get the location of these features at the new frame as our output and perform the same operations between the old and new coordinates as explained in sections \ref{relative_speed} and \ref{trajectory} to get speed $S_{own, x}$ and trajectory $\boldsymbol{T_{own, x}}$ for each feature $x$. The speed and trajectory values are averaged across the set of all $n$ features to obtain the speed and trajectory of our vehicle $S_{own}$ and $\boldsymbol{T_{own}}$. These values are then utilized to compute the absolute speed of other tracked vehicles in the frame as explained ahead. 

\subsubsection{Absolute speed of tracked vehicles in the frame}
\label{tracked_absolute_speed}
We determine the absolute speed of the tracked vehicles by performing the following mathematical calculations. The calculations use the absolute speed $S_{own}$ and trajectory $\boldsymbol{T_{own}}$ of our own vehicle on which the dashcam is mounted (as determined in Section \ref{own_absolute_speed}) along with the direction vector $\boldsymbol{\nu}$ (as obtained from Section \ref{trajectory}) and the relative scaled speed (${S_s}$) (as calculated in Section \ref{relative_speed}) of the tracked vehicle. We calculate the Absolute Speed ($S_{abs}$) of tracked vehicle as shown in \eqref{eq:abs_speed} using the aforementioned values.

\begin{equation}\label{eq:abs_speed}
S_{abs} = \left|\left(S_{own}\times{\frac{\boldsymbol{T_{own}}}{|\boldsymbol{T_{own}}|}}\right) \boldsymbol{+} \left(S_{s}\times{\frac{\boldsymbol{\nu}}{|\boldsymbol{\nu}|}}\right)\right|
\end{equation}

\subsubsection{Determining the change in acceleration of tracked vehicles}
We determine the change in acceleration of the tracked vehicles by using the absolute speed ($S_{abs}$) values (obtained in Section \ref{tracked_absolute_speed}). Then, for a given ${Interval}$ (as explained in Section \ref{relative_speed}), the change in acceleration (${A}$) of the vehicle is calculated from the difference in absolute speed from ${S_{abs}^1}$ to ${S_{abs}^2}$ using \eqref{eq:7}.
\begin{equation}\label{eq:7}
A = \frac{S_{abs}^2 - S_{abs}^1}{\tau \times Interval}
\end{equation}

Change in acceleration (${A}$) serves as a parameter for determining if a collision has occurred.\par

\subsection{Accident Detection}
This part details the procedure of accident detection when tracked vehicles are found to be overlapping as shown in \figurename~\ref{fig:accidentworkflow}. Three new parameters (${\alpha, \beta, \gamma}$) are introduced to determine the anomalies for an accident. The parameters are as follows:
\begin{enumerate}
	\item Anomaly of Acceleration, $\alpha$
	\item Anomaly of Trajectory, $\beta$
	\item Anomaly regarding the change in angle, $\gamma$
\end{enumerate}

At the point when two vehicles are deemed to be intersecting, the acceleration of the vehicles is acquired from their paces caught in the dictionary. The mean acceleration of the vehicles is acquired from 15 frames before the frame at which overlapping condition C1 is met, and the highest acceleration of the automotive is calculated from the 15 frames after C1. The difference in acceleration of the individual vehicles is realised by subtracting the mean acceleration from the highest acceleration, in the period of overlapping condition (C1). The Acceleration Anomaly ($\alpha$) is a score that is characterized to recognize collision dependent on this distinction with a pre-determined set of conditions. This parameter catches the adjustment in a vehicle's speed during a crash consequently allowing the discovery of mishaps from its variation.\par

The Trajectory Anomaly ($\beta$) is calculated from the angle of intersection of the trajectories of vehicles ($\theta$) when the vehicles are considered to be overlapping as per condition C1. 

\begin{enumerate}
	\item If $\theta \in (\theta_L$ $\theta_H)$, $\beta$ is specified from a pre-determined collection of conditions on the magnitude of $\theta$.
	\item Otherwise, $\beta$ is computed from $\theta$ and the separation of the point of intersection of the trajectories from a pre-determined set of conditions.
\end{enumerate}

This parameter is found out by determining the angle ($\theta$) of a vehicle with respect to its own direction over the course of five frames. In case of accident, a vehicle experiences some level of rotation concerning its own axis. By using the adjustment in angles of the trajectories of a vehicle, this degree of rotation can be found permitting the framework to determine the degree to which the vehicle has experienced an orientation change. The Change in Angle Anomaly ($\gamma$) is determined for each vehicle involved in the overlapping criteria depending on a pre-characterized set of conditions.\par 

\begin{figure*}[!ht]
    \centering
	\includegraphics[width=\textwidth, height = 32mm]{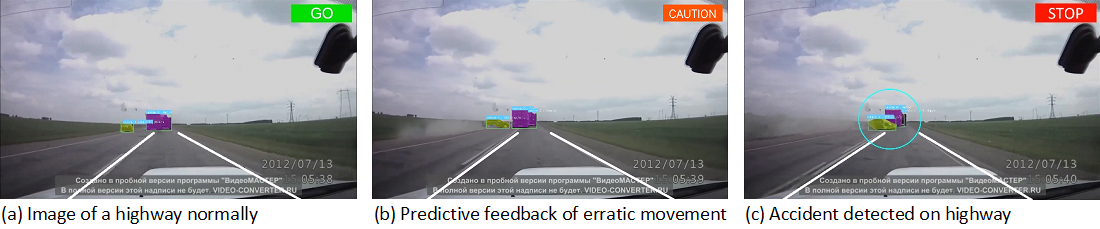}
	\caption{The steps for accident detection on a highway}
	\label{fig:accidents}
\end{figure*}

In the end, all the separately decided anomaly scores are joined with the assistance of a function to decide if a mishap has happened. This function ${f(\alpha, \beta, \gamma)}$ gives weightage to every one of the separate thresholds dependent on their values and produces a score somewhere in the range of ${0}$ and ${1}$. A score which is more than ${0.5}$ informs us that a vehicular mishap has happened else it is disposed of. This is the essential guideline for recognizing an accident and the whole pipeline of distinguishing accidents on real examples can be represented in \figurename~\ref{fig:accidents}.

\subsection{Combination}
To combine the whole approach, some hardware and software-based assumptions are made. Our model presumes that there is a central data collection unit that has a single high-definition camera in front of the driverless vehicle (dashboard-camera) which records real-time video and forwards the visual data to a singular graphical processing unit which extracts relevant actionable information. Video frames are fed into the Mask R-CNN network which performs initial feature extraction and the output is in turn consumed by various task-specific branches. We find that the combined framework is able to perform as well as the individual units when tested on real-time videos. Predictions made in terms of detecting environmental information (detection of lanes, traffic lights, etc.) and generating suggested actions are presented visually over the video frames which can be converted to mechanical actions (turning the vehicle, slowing down, etc.) by wiring the GPU to the electrical circuitry of the vehicle. The details on data collection and results of experiments performed are presented in the following section.

\section{Experimental evaluation}
\label{section3}
All the experiments were conducted on Intel\textregistered\space  Xeon\textregistered\space CPU @ 2.30GHz with NVIDIA Tesla K80 GPU, 12GB VRAM, and 12GB Main Memory (RAM). All programs were written in ${Python - 3.5}$ and utilized ${Keras - 2.2.4}$ and ${Tensorflow - 1.12.0}$. Video processing was done using ${OpenCV 4.0}$.

\subsection{Dataset Used}

The following datasets have been used for testing the Vehicular Accident Detection framework:

\begin{enumerate}
	\item MS COCO Dataset: The Microsoft Common Objects in Context (MS COCO) dataset \cite{DC_51_1} is a well-known dataset that has annotations for instance segmentation and ground truth bounding boxes, which are used to evaluate how well object detection and image segmentation models perform. It contains 91 common objects as classes. Out of the total classes, 82 have 5,000 annotations or more. There are overall 328,000 images with more than 2,500,000 annotated objects. The MS COCO dataset has been used for training the Mask R-CNN framework on multiple object classes including several vehicles such as cars, buses, and trucks among the object classes.
	
	\begin{figure}[!ht]
    	\centering
    	\includegraphics[width=0.95\linewidth, height=0.3\textheight]{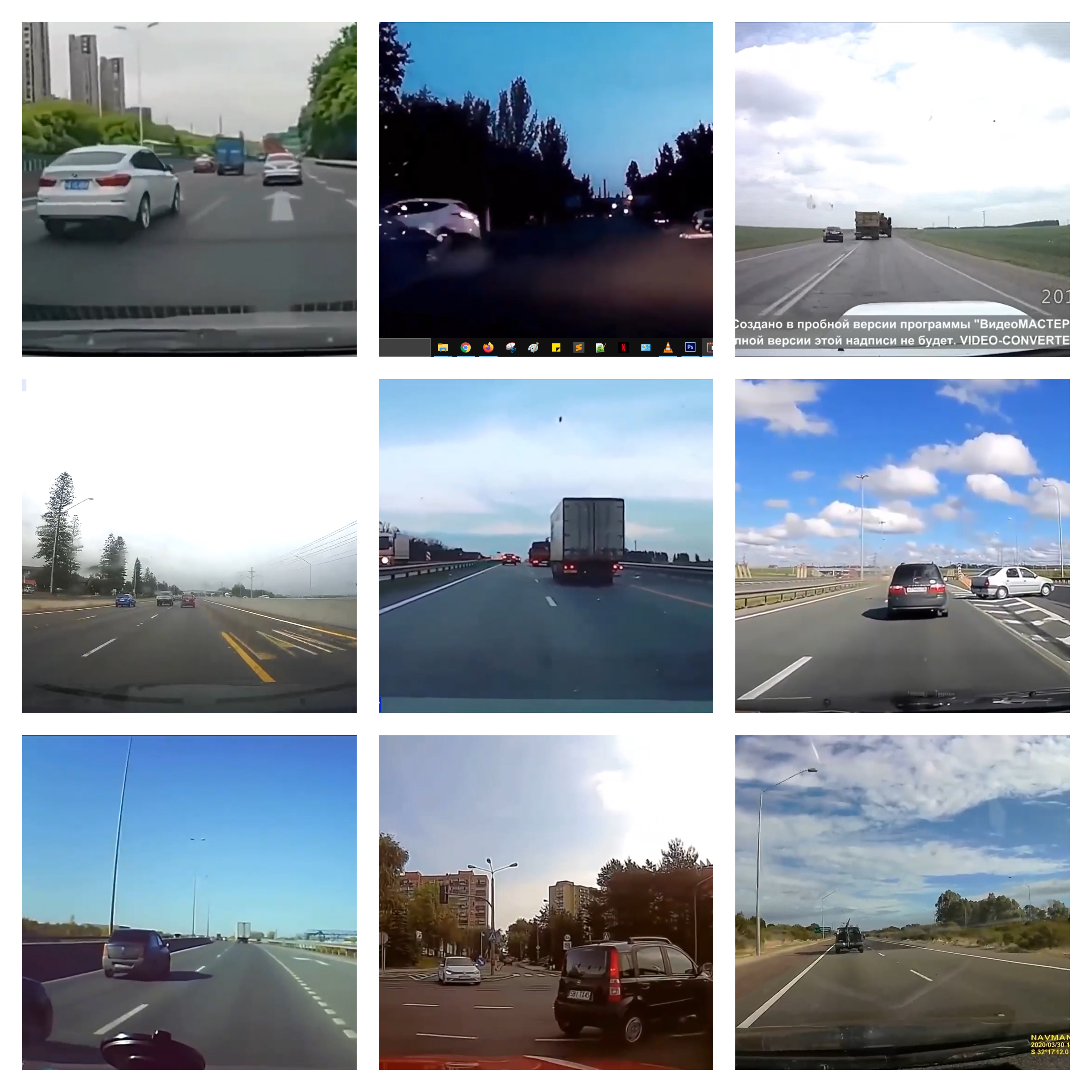}
    	\caption{Screenshots taken from test videos of accidents occurring on roads dataset}
    	\label{fig:accidentcollage}
	\end{figure}
	
	\item Custom dataset for accidents on roads: Accuracy of Mask R-CNN and the overall framework has been evaluated on dash-cam footage of accidents occurring on roads. These videos have been compiled from YouTube. The dash-cam videos have been taken at 30 frames per second and consist of footage from various countries around the world. A handful of dataset images are shown in \figurename~\ref{fig:accidentcollage}.
	
\end{enumerate}

\subsection{Results, Statistics, and Comparison with Existing models}

The performance of the vehicular accident detection framework (as explained in Section \ref{accident_method}) is based on two major criteria i.e., Accident Detection Rate (ADR) and False Alarm Rate (FAR). These two parameters have been used in the majority of the comparative studies and therefore are the most suitable when it comes to comparing with other models. The two parameters are formulated as shown by \eqref{eq:DR} and \eqref{eq:FAR_animal}.

\begin{equation}
\label{eq:DR}
\text{Accident Detection Rate} = \frac{\text{Identified accidents}}{\text{Total accidents}} \times 100\% 
\end{equation}

\begin{equation}
\label{eq:FAR_animal}
\text{FAR} = \frac{\text{Patterns where false alarm occurs}}{\text{Total number of patterns}}\times 100\%
\end{equation}

The existing dashcam accident detection algorithms utilize a reduced total of varying conditions when compared to the videos tested in this framework. Therefore, in contrast to the existing set of works, our approach utilizes a more pragmatic dataset for evaluation, as described in \tablename~\ref{tb:vehicleresults}.

{\renewcommand{\arraystretch}{1.5}%
	\begin{table}[!ht]
		\centering
		\caption{Comparison to other frameworks for Vehicular Accident Detection with proposed approach}
		\begin{tabular}{|c|c|c|}
			\hline
			Approach & ADR (\%) & FAR (\%) \\ \hline
			Sahrawat \emph{et al.}\cite{sahrawat2019improving} & 71.33 & 59.23 \\ \hline
			Chan \emph{et al.}\cite{chan2016anticipating} & 80.00 & 38.45 \\ \hline
			Our Framework (Section \ref{accident_method}) & 79.05 & 34.44
			\\ \hline
		\end{tabular}
		\label{tb:vehicleresults}
	\end{table}
} \quad
The proposed system can recognize vehicular accidents accurately with 79.05\% Accident Detection Rate as per the calculations given in \eqref{eq:DR} and 34.44\% False Alarm Rate as per the calculations given in  \eqref{eq:FAR_animal}. The novelty of our framework is magnified due to the inclusion of a number of parameters for assessing the chances of a mishap which have been recorded under different ambient conditions, including sunlight, night and snow.\par

\section{Conclusion and Future Works}
\label{section4}
\par A novel framework for identifying vehicular impacts is proposed. This system depends on the extraction of information from tracked vehicles. For instance, trajectory intersection, velocity computation, and lane detection are observed through the video feed in a dash-cam. The reliability of our framework is magnified due to the inclusion of a number of parameters for assessing the chances of a mishap. The proposed system can recognize accidents accurately with 79.05\% Accident Detection Rate and 34.44\% False Alarm Rate on the recordings acquired under different ambient conditions.

\par The experimental results are encouraging and showcase the ability of the overall framework. Our work contributes to the development of self-driving car infrastructure in Asian countries like India. However, one compartment in which our system lacks is in the availability of large enough annotated datasets to test and improve the system. Moreover, the system is bottlenecked by the prediction power of the Mask R-CNN framework. Future work in our domain will attempt to fix these issues and incorporate even more support systems.

\section{Acknowledgment}

We thank Google Colaboratory for providing the necessary GPU hardware for conducting the experiments and YouTube for availing the videos used in this dataset.

\bibliographystyle{IEEEtran}
\bibliography{main}
\end{document}